\documentclass[10pt,twocolumn,letterpaper]{article}

\usepackage[pagenumbers]{wacv} 

\usepackage{graphicx}
\usepackage{amsmath}
\usepackage{amssymb}
\usepackage{booktabs}
\usepackage{array}
\usepackage{mwe}
\usepackage{comment}

\usepackage{algorithm}
\usepackage[noend]{algorithmic}

\usepackage[pagebackref,breaklinks,colorlinks]{hyperref}

\usepackage[capitalize]{cleveref}
\crefname{section}{Sec.}{Secs.}
\Crefname{section}{Section}{Sections}
\Crefname{table}{Table}{Tables}
\crefname{table}{Tab.}{Tabs.}

\def\wacvPaperID{11} 
\def\confName{WACV}
\def\confYear{2025}

\begin{document}

\title{Improving the Perturbation-Based Explanation of Deepfake Detectors\\ Through the Use of Adversarially-Generated Samples \thanks{This work was supported by the EU Horizon Europe program under grant agreement 101070190 AI4TRUST.}}

\author{Konstantinos Tsigos \hspace{20mm} Evlampios Apostolidis \hspace{20mm} Vasileios Mezaris\\
Information Technologies Institute - CERTH, Thessaloniki, Greece, 57001\\
{\tt\small \{ktsigos, apostolid, bmezaris\}@iti.gr}
}
\maketitle

\begin{abstract}
   In this paper, we introduce the idea of using adversarially-generated samples of the input images that were classified as deepfakes by a detector, to form perturbation masks for inferring the importance of different input features and produce visual explanations. We generate these samples based on Natural Evolution Strategies, aiming to flip the original deepfake detector's decision and classify these samples as real. We apply this idea to four perturbation-based explanation methods (LIME, SHAP, SOBOL and RISE) and evaluate the performance of the resulting modified methods using a SOTA deepfake detection model, a benchmarking dataset (FaceForensics++) and a corresponding explanation evaluation framework. Our quantitative assessments document the mostly positive contribution of the proposed perturbation approach in the performance of explanation methods. Our qualitative analysis shows the capacity of the modified explanation methods to demarcate the manipulated image regions more accurately, and thus to provide more useful explanations.
\end{abstract}

\section{Introduction}
\label{sec:intro}

\begin{figure*}[t]
    \centering
    \begin{subfigure}{0.16\textwidth}
         \centering
        \frame{\includegraphics[width=\textwidth]{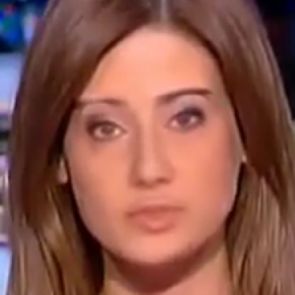}}
        \caption{Original}
    \end{subfigure}
    \hfill
    \begin{subfigure}{0.16\textwidth}
        \centering
        \frame{\includegraphics[width=\textwidth]{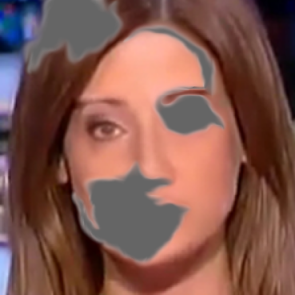}}
        \caption{Occlude}
    \end{subfigure}
    \hfill
    \begin{subfigure}{0.16\textwidth}
        \centering
        \frame{\includegraphics[width=\textwidth]{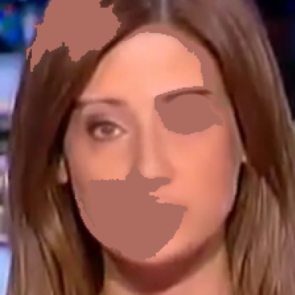}}
        \caption{Replace}
    \end{subfigure}
    \hfill
    \begin{subfigure}{0.16\textwidth}
        \centering
        \frame{\includegraphics[width=\textwidth]{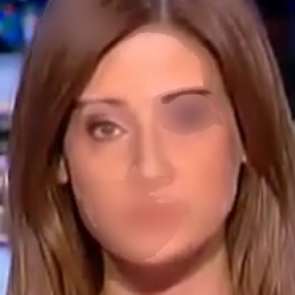}}
        \caption{Blurring}
    \end{subfigure}
    \hfill
    \begin{subfigure}{0.16\textwidth}
        \centering
        \frame{\includegraphics[width=\textwidth]{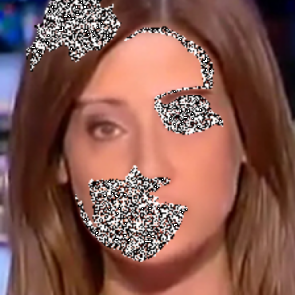}}
        \caption{Gaussian noise}
    \end{subfigure}
    \hfill
    \begin{subfigure}{0.16\textwidth}
        \centering
        \frame{\includegraphics[width=\textwidth]{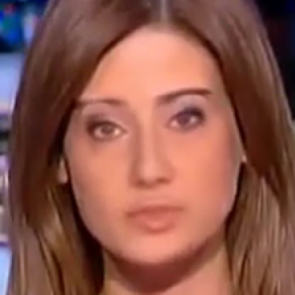}}
        \caption{Adversarial masking}
    \end{subfigure}
    \caption{The original image (a) and the perturbed instances of it after: occluding certain segments (b), replacing them with fixed values (c), blurring them (d), adding Gaussian noise (e), and (f) replacing them with the corresponding segments of an adversarially-generated sample of the input image (Original face image (a) source: FaceForensics++ dataset).}
   \label{fig:perturbations}
\end{figure*}

Over the last decade, there is an ongoing advancement and use of AI technologies for tackling a variety of tasks, such as natural language processing \cite{LAURIOLA2022443} and text classification \cite{10.1145/3495162}, audio classification \cite{10258355} and speaker diarization \cite{PARK2022101317}, as well as video understanding \cite{madan2024} and summarization \cite{9594911}. Nevertheless, the ``black-box'' nature of AI models raises concerns about their use in several sensitive (in terms of human safety) applications and domains, such as autonomous cars, cybersecurity and medical decision-making \cite{Yang2023,MERSHA2024128111}. Transparency, accountability and explainability are crucial for building trust in AI, and over the last years there is a shift from just using the power of AI, to understanding and interpreting how AI has made a decision.

Several methods have been proposed for explaining the output of AI models, \eg \cite{10.1145/2939672.2939778,10.5555/3295222.3295230,fel2021sobol,DBLP:conf/bmvc/PetsiukDS18,8354201,10539635,9093360}. According to \cite{MERSHA2024128111}, these methods can be categorized based on their scope (local or global), the stage they are implemented (during (``ante-hoc'') or after (``post-hoc'') the training), and the methodology (\eg perturbation-based, gradient-based). In this paper, we focus on ``post-hoc'' perturbation-based methods that do not require any access to the model's internal structure and design (\ie are model-agnostic). These methods perturb the input data, observe the changes in the model's output, and analyze the extent to which the output is affected by each perturbation. Through a series of perturbations and corresponding outputs, they are able to infer the most important features for the model's decisions \cite{10.1145/2939672.2939778}.

Various types of perturbations have been described in the literature on explainable image/video classifiers, such as the removal of input features, their masking with fixed or random values, and their replacement by blurred patches or Gaussian noise \cite{BAI2021108102,MERSHA2024128111}. Such perturbations are meaningful when explaining the output of models for \eg image/video annotation. However, they are less suitable when the aim is to explain the decision of deepfake image/video detectors. \cref{fig:perturbations} shows perturbed instances of an example deepfake image (\cref{fig:perturbations}a) after occluding certain segments of it (\cref{fig:perturbations}b), replacing them with a fixed value (\cref{fig:perturbations}c), blurring them (\cref{fig:perturbations}d), or adding Gaussian noise (\cref{fig:perturbations}e). As can be seen, these perturbations result in images that are clearly visually dissimilar from the original one. As documented in \cite{10.1007/978-3-031-09037-0_8} and discussed also in \cite{GOWRISANKAR2024103684}, such perturbations may produce images that do not lie within the distribution of the data used for training the deepfake detector, giving rise to the out-of-distribution (OOD) issue and leading to unexpected model behavior. As a consequence, the applied explanation method cannot accurately detect whether the observed change in the deepfake detector's output is associated with the removal/modification of important input features or with the shift in the data distribution. 

To overcome the aforementioned weakness of existing perturbation-based explanation methods, in this paper we propose the use of an adversarially-generated sample of the input deepfake image that flips the detector's decision (thus it is classified as ``real''), to form perturbation masks for inferring the importance of different features. More specifically, inspired by the evaluation framework of \cite{10.1145/3643491.3660292}, we iteratively perform adversarial attacks in the input image based on Natural Evolution Strategies (NES) \cite{wierstra14a}, until the generated sample is classified as ``real'' or a maximum number of iterations $M$ is reached. During this iterative process, we make sure that only a small magnitude of noise is added to the input image in order to generate samples that remain visually imperceptible from it. Consequently, when the finally generated adversarial image is used to form perturbation masks (as depicted in the lower part of \cref{fig:perturbations2}), the perturbed instances of the input image after the so-called adversarial masking are also visually-similar with this image, as shown in \cref{fig:perturbations}f, avoiding OOD issues. Given that the adversarial images were generated in order to flip the deepfake detector's decision (\ie to make the detector classify them as ``real''), we argue that their use to form perturbation masks for the input deepfake images will allow explanation methods to infer the importance of different features more effectively, and produce more meaningful and accurate visual explanations. Our contributions are as follows:
\begin{itemize}
    \item We propose a new perturbation approach that is based on the use of adversarially-generated samples of the detected deepfake images, to form perturbation masks for inferring the importance of different input features.
    \item We extend and adapt four SOTA explanation methods (RISE \cite{DBLP:conf/bmvc/PetsiukDS18}, SHAP \cite{10.5555/3295222.3295230}, LIME \cite{10.1145/2939672.2939778}, SOBOL \cite{fel2021sobol}) in order to integrate the proposed perturbation approach.
    \item We evaluate the performance of the modified explanation methods quantitatively (based on \cite{10.1145/3643491.3660292}) and qualitatively, and document the gains of integrating the proposed perturbation approach.
\end{itemize}

\section{Related Work}

\begin{figure*}[t]
\centering
\includegraphics[width=.96\textwidth]{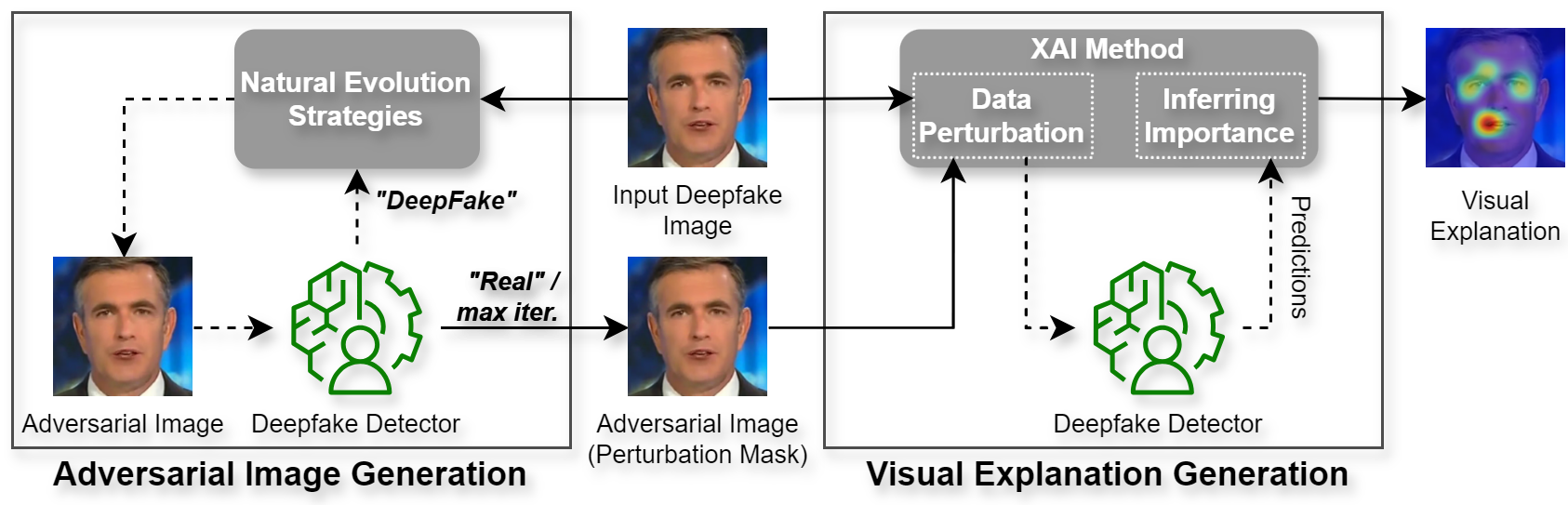}
\caption{The processing pipeline of the proposed explanation approach; dashed lines indicate iterative processes. (Input Deepfake Image source: FaceForensics++ dataset).}
\label{fig:pipeline}
\end{figure*}

Despite the ongoing interest in developing AI-based models for detecting deepfake images and videos, the explanation of the output of these models has been investigated only to a small extent, thus far. In one of the first attempts towards explaining a deepfake detector's decision, Malolan \etal \cite{9092227}, examined the use of the LIME \cite{10.1145/2939672.2939778} and LRP \cite{10.1371/journal.pone.0130140} methods to produce visual explanations (in the form of heatmaps) about the output of an XceptionNet-based \cite{8099678} detector that was trained on a subset of the FaceForensics++ dataset \cite{9010912}. Pino \etal \cite{DBLP:journals/corr/abs-2105-05902} employed adaptations of SHAP \cite{10.5555/3295222.3295230}, Grad-CAM \cite{8237336} and self-attention methods, to explain the output of deepfake detectors (based on EfficientNet-B4 and B7 \cite{tan2020efficientnet}) that were trained on the DFDC dataset \cite{DFDC2020}. Silva \etal \cite{SILVA2022100217}, applied the Grad-CAM method \cite{8237336} and focused on the computed gradients for the attention map to spot the image regions that influence the most the predictions of an ensemble of CNNs (XceptionNet \cite{8099678}, EfficientNet-B3 \cite{DBLP:conf/icml/TanL19}) and attention-based models for deepfake detection. Xu \etal \cite{9707568}, utilized learned features to explain the detection performance of a trained linear deepfake detector on the FaceForensics++ dataset \cite{9010912}, with the help of heatmap visualizations and uniform manifold approximation and projection. Jayakumar \etal \cite{9993294}, employed the Anchors \cite{10.5555/3504035.3504222} and LIME \cite{10.1145/2939672.2939778} methods to produce visual explanations for a trained model (based on EfficientNet-B0 \cite{DBLP:conf/icml/TanL19}) that detects five different types of deepfakes. Aghasanli \etal \cite{10350382}, used SVM and xDNN \cite{ANGELOV2020185} classifiers to understand the behavior of a Transformer-based deepfake detector. Finally, Jia \etal \cite{Jia_2024_CVPR} examined the capacity of multimodal LLMs in detecting deepfakes and providing textual explanations about their decisions.

In terms of evaluation, most of the aforementioned works evaluate the produced explanations qualitatively using a small set of samples \cite{9092227,9707568,SILVA2022100217,9993294,10350382,Jia_2024_CVPR}. Towards a quantitative, and thus more objective, evaluation, Pino \etal \cite{DBLP:journals/corr/abs-2105-05902} used tailored metrics that relate to low-level features of the obtained visual explanations, such as inter/intra-frame consistency, variance and centredness. Gowrisankar \etal \cite{GOWRISANKAR2024103684} described a framework that applies a number of adversarial attacks, adding noise in regions of a deepfake image that correspond to the spotted salient regions after explaining the (correct) classification of its real counterpart, and evaluates the performance of an explanation method based on the observed drop in the detector's accuracy. Building on \cite{GOWRISANKAR2024103684}, Tsigos \etal \cite{10.1145/3643491.3660292} proposed a simpler evaluation framework, which uses the produced explanation after detecting a deepfake image and does not require access to its original counterpart. Finally, the ROAD framework \cite{pmlr-v162-rong22a} for evaluating attribution-based explanation methods for image classifiers, applies a similar noise-based imputation strategy. Nevertheless, the efficiency of the applied imputation has not been investigated thus far on deepfake classifiers. 

Our paper is most closely related with works that utilize \cite{9092227,9993294} or evaluate \cite{GOWRISANKAR2024103684,10.1145/3643491.3660292} perturbation-based explanation methods. Nevertheless, contrary to these works, we propose the use of adversarially-generated perturbation masks and investigate their impact on inferring the importance of input features using modified versions of four SOTA perturbation-based explanation methods (RISE \cite{DBLP:conf/bmvc/PetsiukDS18}, SHAP \cite{10.5555/3295222.3295230}, LIME \cite{10.1145/2939672.2939778}, SOBOL \cite{fel2021sobol}). Moreover, in contrast to \cite{9092227,9707568,SILVA2022100217,9993294,10350382}, that evaluate explanations only qualitatively using a small set of samples, we assess the performance of explanation methods also using a quantitative evaluation framework.

\section{Proposed Approach}

\begin{figure*}[t]
\centering
\includegraphics[width=0.77\textwidth]{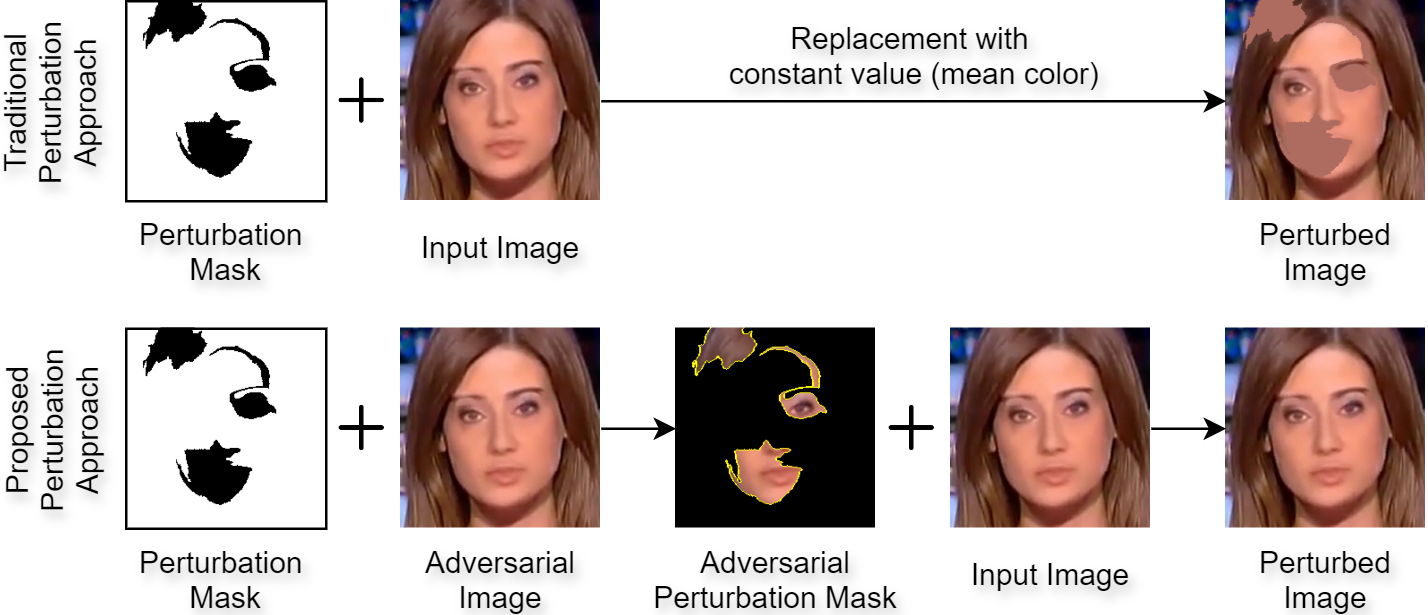}
\caption{Top: Example of a traditional perturbation approach. Bottom: Illustration of the proposed perturbation approach that uses the adversarially-generated sample of the input image.}
\label{fig:perturbations2}
\end{figure*}

\begin{algorithm}[t]
\caption{Adversarial image generation}\label{alg:advimgen}
\begin{algorithmic}[1]
\renewcommand{\algorithmicrequire}{\textbf{Parameters:}}
{\REQUIRE Search variance $\sigma$, Number of samples $\mathit{n}$, Image width/height $\mathit{D}$, Maximum number of iterations $M$, Maximum distortion $\delta$, Learning rate $\alpha$, Gaussian noise generator $\mathcal{N}(mean,covar.)$, Identity matrix $\boldsymbol{I}$}
\renewcommand{\algorithmicrequire}{\textbf{Input:}}
{\REQUIRE Input image $\boldsymbol{x}$, Deepfake detector $\mathcal{F}(\boldsymbol{x})$ with $\mathcal{F}(\boldsymbol{x})_{real}$ being the probability to classify $\boldsymbol{x}$ as ``real''}
\renewcommand{\algorithmicrequire}{\textbf{Output:}}
{\ENSURE Adversarial image $\boldsymbol{x_{adv}}$}
\STATE $\boldsymbol{x_{adv}} = \boldsymbol{x}$,
\FOR{$i=1 \to M$}
    \STATE \textbf{if} $F(\boldsymbol{x_{adv}})$ is $real$ \textbf{then} return $\boldsymbol{x_{adv}}$
    \STATE $\boldsymbol{g} = 0_{D,D}$ \hspace{2mm}\textit{\# Matrix that stores gradients}
    \FOR{$j=1 \to n$}
        \STATE $\boldsymbol{u}_{j} = \mathcal{N}(0_{D},\boldsymbol{I}_{D,D})$ \hspace{2mm}\textit{\# Random noise}
        \STATE $\boldsymbol{g} = \boldsymbol{g} + \mathcal{F}(\boldsymbol{x_{adv}} + \sigma \cdot\boldsymbol{u}_{j})_{real} \cdot \boldsymbol{u}_{j}$
        \STATE $\boldsymbol{g} = \boldsymbol{g} - \mathcal{F}(\boldsymbol{x_{adv}} - \sigma \cdot \boldsymbol{\boldsymbol{u}_{j}})_{real} \cdot \boldsymbol{u}_{j}$
    \ENDFOR
    \STATE $\boldsymbol{g} = \frac{1}{2n\sigma}\boldsymbol{g}$
    \STATE $\boldsymbol{x_{adv}} = \boldsymbol{x_{adv}} + clip_{\delta}(\alpha \cdot sign(\boldsymbol{g}))$
\ENDFOR
\STATE return $\boldsymbol{x_{adv}}$
\end{algorithmic}
\end{algorithm}

The proposed approach aims to produce a visual explanation after a deepfake detector classifies an input image as a deepfake, providing clues about regions of the image that were found to be manipulated. The processing pipeline, shown in \cref{fig:pipeline}, contains an adversarial image generation step and a visual explanation generation step. The former aims to create an adversarial sample of the input image that can fool the detector to classify it as ``real'', and the latter uses this sample to form perturbation masks for inferring the importance of different parts of the input image and generate the visual explanation (\eg in the form of a heatmap). 

Assuming a trained deepfake detector and an input image that has been classified by it as a deepfake, the adversarial image generation is performed through an iterative process (presented in \cref{alg:advimgen}) that stops when the deepfake detector is fooled to classify the generated adversarial sample as ``real'' or when a maximum number of iterations $M$ is reached. During each iteration, this process generates a number of samples of the input image using NES \cite{wierstra14a}. The employed NES technique, is a black-box optimization approach (thus, it does not require access to the internal structure of the detector) that was used in \cite{9423202} for adversarial deepfake generation. It estimates the gradient of the detector's output with respect to the addition of small random Gaussian noise to the entire image and tries to find a minimal change in the pixels' values that will cause the detector to mis-classify the generated image as ``real''. To reduce variance and make the pixel-level gradient estimation more robust, this process is repeated for $\mathit{n}$ noisy samples per iteration. Moreover, to make sure that the generated samples remain visually imperceptible from the input image and avoid OOD issues, the applied technique uses a specified search variance $\sigma$ that controls the amount of random Gaussian noise added to the image when exploring the space of possible perturbations. Finally, a maximum distortion $\delta$ defines the maximum allowed change in pixel values, and a learning rate $\alpha$ specifies the step size for updating the adversarial image after each iteration.

The deepfake detector, along with the generated adversarial image and the input image, are then used during the visual explanation generation step. In particular, both images are given as input to the explanation method, which uses the former one to form perturbation masks for the latter one. So, contrary to traditional perturbation-based explanation methods that replace regions of the input image \eg using a constant value (as depicted in the example in the upper part of \cref{fig:perturbations2}), our proposed perturbation approach replaces these regions by using the corresponding parts of the adversarially-generated image (as shown in the example in the lower part of \cref{fig:perturbations2}). This process is repeated several times (depending on the explanation method), resulting in an equivalent number of perturbed images that are analyzed by the deepfake detector. The applied perturbations and the corresponding predictions of the deepfake detector, are finally taken into account by the explanation method to infer the importance of different input features and create the visual explanation. 

\section{Experiments}

\subsection{Explanation methods}

We considered the following explanation methods:
\begin{itemize}
    \item \textbf{LIME} \cite{10.1145/2939672.2939778} creates visual explanations by replacing portions of the input image with the mean pixel value of the image (as in \cref{fig:perturbations}c) to assess their influence on the model's output. It approximates the model's behavior locally by fitting the perturbation data and the corresponding model's outputs into a simpler model (\eg a linear regressor).
    \item \textbf{SHAP} \cite{10.5555/3295222.3295230} leverages the Shapley values from game theory. It constructs an additive feature attribution model that attributes an effect to each input feature and sums the effects, i.e., SHAP values, as a local approximation of the output.
    \item \textbf{SOBOL} \cite{fel2021sobol} generates a set of real-valued masks and applies them to an input image through blurring-based perturbations (\ie as in \cref{fig:perturbations}d). By analyzing the relationship between masks and the corresponding model's prediction, it estimates the order of Sobol' indices and creates an explanation by highlighting each region's importance.
    \item \textbf{RISE} \cite{DBLP:conf/bmvc/PetsiukDS18} produces a set of binary masks to occlude regions of the input image (as in \cref{fig:perturbations}b) and produce a set of perturbations. Then, it feeds these perturbations to the model, gets its predictions, uses them to weight the corresponding masks, and creates the explanation by aggregating the weighted masks together.
\end{itemize}

\subsection{Dataset and evaluation protocol}

We ran experiments on the FaceForensics++ dataset \cite{9010912}, which includes $1000$ original videos and $4000$ fake videos. The fake videos are equally divided into four different classes ($1000$ per class) according to the type of the applied AI-based manipulation for generating them. More specifically, the videos of the ``FaceSwap'' (FS) class were produced by transferring the face region from a source to a target video. The videos of the ``DeepFakes'' (DF) class were generated by replacing a face in a target sequence with a face in a source video or image collection. The videos of the ``Face2Face'' (F2F) class were created by transferring the expressions of a source to a target video while maintaining the identity of the target person. Finally, the videos of the ``NeuralTextures'' (NT) class were obtained after modifying the facial expressions corresponding to the mouth region. The dataset is split into training, validation, and test sets, containing $720$, $140$ and $140$ videos, respectively. 

Quantitative performance evaluations were made using the evaluation framework and data of \cite{10.1145/3643491.3660292}, downloaded from\footnote{\url{https://github.com/IDT-ITI/XAI-Deepfakes}\label{url}}. These data comprise four sets of $1270$ images (one set per different class of the test set of the dataset). Assuming an image classified as a deepfake and the produced explanation, the used framework assesses the performance of the applied explanation method by examining the extent to which the image regions that were found as the most important ones, can be used to flip the deepfake detector's decision. For this, after segmenting the input image into super-pixel segments based on the SLIC algorithm \cite{6205760}, the evaluation framework quantifies the importance of each segment by averaging the computed explanation scores at the segment-level. Then, it creates variants of the input image by performing adversarial attacks on the image regions corresponding to the top-k scoring segments using NES, and repeats this process until the detector classifies the generated variant as ``real'' or a maximum number of iterations is reached. We should note that the applied adversarial image generation process during the evaluation process is different from the one performed when generating the adversarial image that is subsequently used for producing the visual explanation. The former is applied locally to the input image, affecting only the top-k segments singled out by the explanation method; the latter is performed globally to the input image, affecting all its pixels. Therefore, there is no overlap in the visual explanation production and evaluation processes, that could put our experimental findings in doubt.

The performance of each explanation method is quantified using two measures: the \textbf{(drop in) accuracy} of the deepfake detector on the adversarially-generated images after affecting the top-1, top-2 and top-3 scoring segments of the input images by the explanation method, and the \textbf{sufficiency} of the explanation methods to spot the most important image regions, measured by calculating the difference in the detector's output after performing adversarial attacks to the top-1, top-2 and top-3 scoring segments of the image. Both measures range in $[0,1]$. Since a larger drop in accuracy is anticipated for explanation methods that spot the most important regions of the input image for the deepfake detector, the lower the accuracy scores, the higher the ability of the explanation method to spot these regions. On the contrary, high sufficiency scores indicate that the top-k scoring segments by the explanation method have high impact to the deepfake detector's decision, and thus the produced visual explanation exhibits high sufficiency.

\subsection{Implementation details}

\begin{table*}[t]
\centering
{\small{
\begin{tabular}{|l|ccc|ccc|ccc|ccc|}
\hline
                  & \multicolumn{3}{c|}{DF}                                                                                & \multicolumn{3}{c|}{F2F}                                                                         & \multicolumn{3}{c|}{FS}                                                                                     & \multicolumn{3}{c|}{NT}                                                                                \\ \hline
Original & \multicolumn{3}{c|}{0.978}                                                                             & \multicolumn{3}{c|}{0.977}                                                                       & \multicolumn{3}{c|}{0.982}                                                                                  & \multicolumn{3}{c|}{0.924}                                                                             \\ \hline
                  & \multicolumn{1}{c|}{Top 1}                & \multicolumn{1}{c|}{Top 2}                & Top 3          & \multicolumn{1}{c|}{Top 1}          & \multicolumn{1}{c|}{Top 2}          & Top 3                & \multicolumn{1}{c|}{Top 1}               & \multicolumn{1}{c|}{Top 2}                & Top 3                & \multicolumn{1}{c|}{Top 1}          & \multicolumn{1}{c|}{Top 2}                & Top 3                \\ \hline
LIME              & \multicolumn{1}{c|}{0.735}                & \multicolumn{1}{c|}{0.440}                 & {\underline {0.245}}    & \multicolumn{1}{c|}{\underline {0.803}}    & \multicolumn{1}{c|}{\underline {0.633}}    & 0.484                & \multicolumn{1}{c|}{0.864}               & \multicolumn{1}{c|}{0.698}                & 0.559                & \multicolumn{1}{c|}{0.579}          & \multicolumn{1}{c|}{0.340}                 & 0.197                \\
LIME$_{adv}$  & \multicolumn{1}{c|}{ \textbf{0.673}} & \multicolumn{1}{c|}{\textbf{0.363}} & \textbf{0.205} & \multicolumn{1}{c|}{\textbf{0.784}} & \multicolumn{1}{c|}{\textbf{0.568}} & \textbf{0.392}       & \multicolumn{1}{c|}{\textbf{0.790}} & \multicolumn{1}{c|}{\textbf{0.538}} & {\textbf{0.378}} & \multicolumn{1}{c|}{\textbf{0.411}} & \multicolumn{1}{c|}{\textbf{0.155}} & {\textbf{0.056}} \\ \hline
SHAP              & \multicolumn{1}{c|}{0.813}                & \multicolumn{1}{c|}{0.609}                & {0.450}     & \multicolumn{1}{c|}{0.846}    & \multicolumn{1}{c|}{0.739}    & 0.637         & \multicolumn{1}{c|}{0.876}               & \multicolumn{1}{c|}{0.702}                & 0.543                & \multicolumn{1}{c|}{0.686}          & \multicolumn{1}{c|}{0.497}                & 0.344                \\
SHAP$_{adv}$  & \multicolumn{1}{c|}{0.706}       & \multicolumn{1}{c|}{\underline{0.434}} & 0.249 & \multicolumn{1}{c|}{0.811} & \multicolumn{1}{c|}{0.647} & {\underline {0.482}} & \multicolumn{1}{c|}{\underline {0.820}} & \multicolumn{1}{c|}{\underline {0.611}} & {\underline {0.430}}  & \multicolumn{1}{c|}{0.485} & \multicolumn{1}{c|}{\underline {0.238}} & {\underline {0.129}} \\ \hline
SOBOL             & \multicolumn{1}{c|}{0.750}                 & \multicolumn{1}{c|}{0.591}                & 0.490           & \multicolumn{1}{c|}{0.816}          & \multicolumn{1}{c|}{0.653}          & 0.512                & \multicolumn{1}{c|}{0.874}               & \multicolumn{1}{c|}{0.703}                & 0.574                & \multicolumn{1}{c|}{0.621}          & \multicolumn{1}{c|}{0.417}                & 0.313                \\
SOBOL$_{adv}$ & \multicolumn{1}{c|}{\underline {0.696}}          & \multicolumn{1}{c|}{0.507}                & 0.383          & \multicolumn{1}{c|}{0.828}          & \multicolumn{1}{c|}{0.680}           & 0.555                & \multicolumn{1}{c|}{0.831}               & \multicolumn{1}{c|}{0.627}                & 0.506                & \multicolumn{1}{c|}{\underline {0.478}}    & \multicolumn{1}{c|}{0.252}                & 0.152                \\ \hline
RISE              & \multicolumn{1}{c|}{0.877}                & \multicolumn{1}{c|}{0.766}                & 0.686          & \multicolumn{1}{c|}{0.843}          & \multicolumn{1}{c|}{0.710}           & 0.622                & \multicolumn{1}{c|}{0.896}               & \multicolumn{1}{c|}{0.809}                & 0.734                & \multicolumn{1}{c|}{0.783}          & \multicolumn{1}{c|}{0.637}                & 0.513                \\
RISE$_{adv}$  & \multicolumn{1}{c|}{0.769}                & \multicolumn{1}{c|}{0.679}                & 0.618          & \multicolumn{1}{c|}{0.876}          & \multicolumn{1}{c|}{0.824}          & 0.784                & \multicolumn{1}{c|}{0.917}               & \multicolumn{1}{c|}{0.867}                & 0.830                 & \multicolumn{1}{c|}{0.515}          & \multicolumn{1}{c|}{0.378}                & 0.322                \\ \hline
\end{tabular}
}}
\caption{Accuracy of the deepfake detector for the different types of fakes (DF, F2F, FS, NT) in the FaceForensics++ dataset, on the original images (second row) and on variants of them after performing adversarial attacks on the image regions corresponding to the top-1, top-2 and top-3 scoring segments based on the different explanation methods. Best (lowest) scores in bold and second best scores underlined.}
\label{tab:accuracy}
\end{table*}

\begin{figure*}[t]
    \centering
    \begin{subfigure}{0.49\textwidth}
        \centering
        \frame{\includegraphics[width=\textwidth]{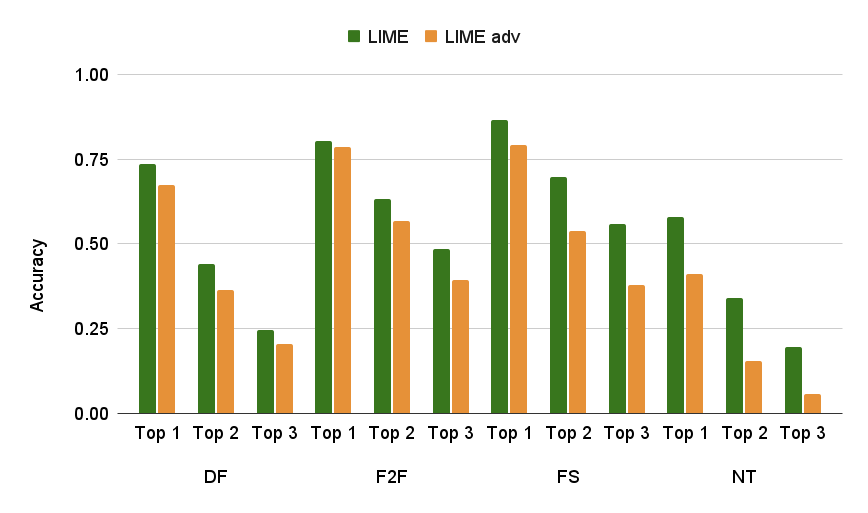}}
        \caption{LIME}
    \end{subfigure}
    \hspace{1mm}
    \begin{subfigure}{0.49\textwidth}
        \centering
        \frame{\includegraphics[width=\textwidth]{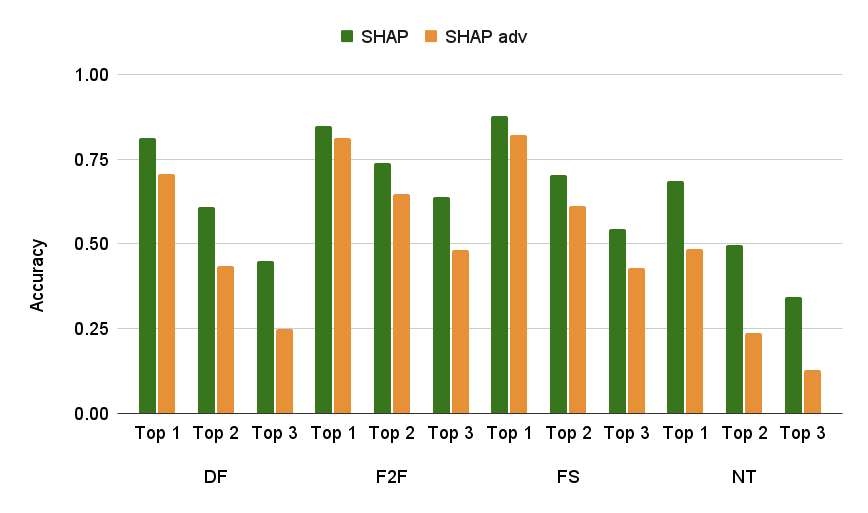}}
        \caption{SHAP}
    \end{subfigure}
    \vskip\baselineskip
    \vspace{-3mm}
    \begin{subfigure}{0.49\textwidth}
        \centering
        \frame{\includegraphics[width=\textwidth]{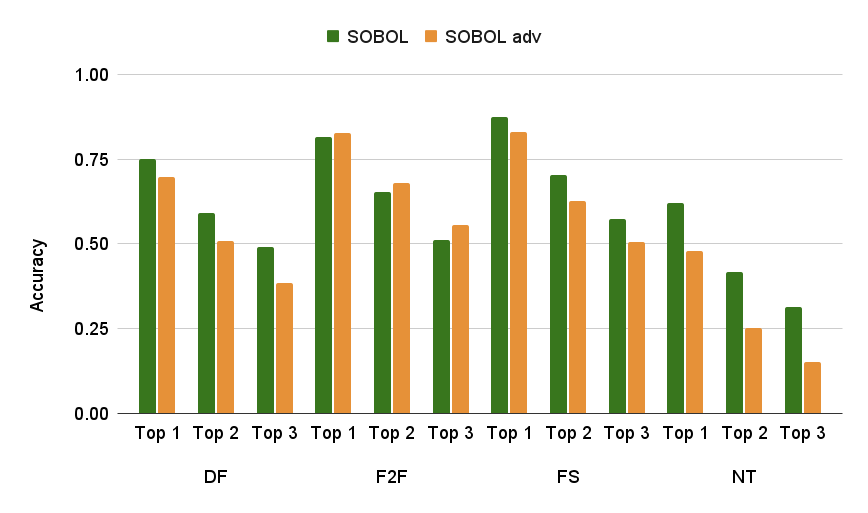}}
        \caption{SOBOL}
    \end{subfigure} 
    \hspace{1mm}
    \begin{subfigure}{0.49\textwidth}
         \centering
        \frame{\includegraphics[width=\textwidth]{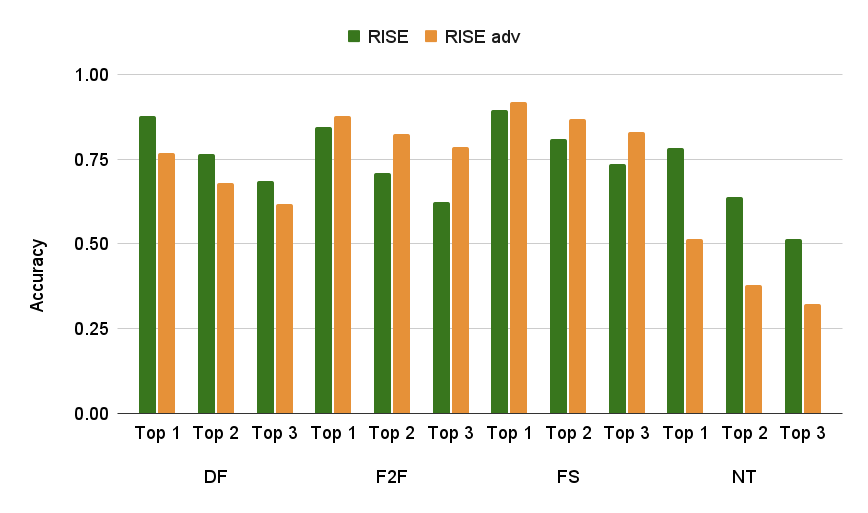}}
        \caption{RISE}
    \end{subfigure}
    \caption{The observed detection accuracy for the different explanation methods and their modified versions. The lower the accuracy, the higher the ability of the explanation method to spot the most important image regions for the deepfake detector's decisions.}
    \vspace{-2mm}
   \label{fig:accuracy_graphs}
\end{figure*}

We used the pretrained deepfake detection model of \cite{10.1145/3643491.3660292}, downloaded from\footref{url}. The deepfake detection model relies on the second version of the EfficientNet architecture \cite{DBLP:conf/icml/TanL21}, that was pretrained on ImageNet 1K and fine-tuned on FaceForensics++, as described in \cite{10.1145/3643491.3660292}. The generation of the visual explanations was based on the following settings:
\begin{itemize}
    \item \textbf{LIME}: set the number of perturbations equal to 2K and used SLIC with number of segments equal to $50$; used python package from: \url{https://pypi.org/project/lime}
    \item \textbf{SHAP}: set the number of evaluations equal to 2K and used a blurring mask with kernel size equal to $128$, when applying the original data perturbation approach; used python package from: \url{https://pypi.org/project/shap}
    \item \textbf{SOBOL}: set the grid size equal to $8$ and the number of design equal to $32$, and kept all other parameters with default values; used implementation from \url{https://github.com/fel-thomas/Sobol-Attribution-Method}
    \item \textbf{RISE}: set the number of masks equal to 4K and kept all other parameters with default values; used implementation from: \url{https://github.com/yiskw713/RISE}
\end{itemize}

With respect to NES, similarly to \cite{GOWRISANKAR2024103684} we set the learning rate $\alpha$ equal to $1/255$, the maximum distortion $\delta$ equal to $16/255$, the search variance $\sigma$ equal to $0.001$, and the maximum number of iterations $M$ during the evaluation stage equal to $50$. Following \cite{10.1145/3643491.3660292}, we set the number of samples $n$ equal to $40$. Finally, the maximum number of iterations during the explanation production step was set equal to $80$. Experiments were carried out on a NVIDIA RTX 4090 GPU cards. Code and model are available at: \url{https://github.com/IDT-ITI/Adv-XAI-Deepfakes}

\subsection{Quantitative results}

\begin{table*}[t]
\centering
{\small{
\begin{tabular}{|l|ccc|ccc|ccc|ccc|}
\hline
                  & \multicolumn{3}{c|}{DF}                                                                    & \multicolumn{3}{c|}{F2F}                                                                   & \multicolumn{3}{c|}{FS}                                                                    & \multicolumn{3}{c|}{NT}                                                                    \\ \cline{2-13}
                  & \multicolumn{1}{c|}{Top 1}          & \multicolumn{1}{c|}{Top 2}          & Top 3          & \multicolumn{1}{c|}{Top 1}          & \multicolumn{1}{c|}{Top 2}          & Top 3          & \multicolumn{1}{c|}{Top 1}          & \multicolumn{1}{c|}{Top 2}          & Top 3          & \multicolumn{1}{c|}{Top 1}          & \multicolumn{1}{c|}{Top 2}          & Top 3          \\ \hline
LIME              & \multicolumn{1}{c|}{0.195}          & \multicolumn{1}{c|}{0.408}          & {\underline {0.539}}    & \multicolumn{1}{c|}{\underline {0.121}}    & \multicolumn{1}{c|}{\underline {0.238}}    & {\underline {0.334}}    & \multicolumn{1}{c|}{0.087}          & \multicolumn{1}{c|}{0.189}          & 0.262          & \multicolumn{1}{c|}{0.233}          & \multicolumn{1}{c|}{0.363}          & 0.431          \\
LIME$_{adv}$  & \multicolumn{1}{c|}{\textbf{0.252}} & \multicolumn{1}{c|}{\textbf{0.464}} & \textbf{0.564} & \multicolumn{1}{c|}{\textbf{0.136}} & \multicolumn{1}{c|}{\textbf{0.276}} & \textbf{0.379} & \multicolumn{1}{c|}{\textbf{0.134}} & \multicolumn{1}{c|}{\textbf{0.278}} & \textbf{0.367} & \multicolumn{1}{c|}{\textbf{0.326}} & \multicolumn{1}{c|}{\textbf{0.455}} & \textbf{0.501} \\ \hline
SHAP              & \multicolumn{1}{c|}{0.137}          & \multicolumn{1}{c|}{0.300}          & 0.402          & \multicolumn{1}{c|}{0.092}          & \multicolumn{1}{c|}{0.158}          & 0.222          & \multicolumn{1}{c|}{0.073}          & \multicolumn{1}{c|}{0.181}          & 0.269          & \multicolumn{1}{c|}{0.167}          & \multicolumn{1}{c|}{0.282}          & 0.357          \\
SHAP$_{adv}$  & \multicolumn{1}{c|}{\underline {0.215}}    & \multicolumn{1}{c|}{\underline {0.423}}    & 0.534          & \multicolumn{1}{c|}{0.112}          & \multicolumn{1}{c|}{0.224}          & 0.324          & \multicolumn{1}{c|}{\underline {0.114}}    & \multicolumn{1}{c|}{\underline {0.240}}    & {\underline {0.334}}    & \multicolumn{1}{c|}{0.280}          & \multicolumn{1}{c|}{\underline {0.405}}    & {\underline {0.463}}    \\ \hline
SOBOL             & \multicolumn{1}{c|}{0.166}          & \multicolumn{1}{c|}{0.277}          & 0.352          & \multicolumn{1}{c|}{0.108}          & \multicolumn{1}{c|}{0.212}          & 0.296          & \multicolumn{1}{c|}{0.078}          & \multicolumn{1}{c|}{0.180}          & 0.259          & \multicolumn{1}{c|}{0.198}          & \multicolumn{1}{c|}{0.302}          & 0.362          \\
SOBOL$_{adv}$ & \multicolumn{1}{c|}{\underline {0.215}}    & \multicolumn{1}{c|}{0.349}          & 0.426          & \multicolumn{1}{c|}{0.105}          & \multicolumn{1}{c|}{0.201}          & 0.270           & \multicolumn{1}{c|}{0.110}          & \multicolumn{1}{c|}{0.221}          & 0.286          & \multicolumn{1}{c|}{\underline {0.282}}    & \multicolumn{1}{c|}{0.393}          & 0.443          \\ \hline
RISE              & \multicolumn{1}{c|}{0.087}          & \multicolumn{1}{c|}{0.162}          & 0.219          & \multicolumn{1}{c|}{0.091}          & \multicolumn{1}{c|}{0.173}          & 0.223          & \multicolumn{1}{c|}{0.060}          & \multicolumn{1}{c|}{0.114}          & 0.157          & \multicolumn{1}{c|}{0.115}          & \multicolumn{1}{c|}{0.204}          & 0.273          \\ 
RISE$_{adv}$  & \multicolumn{1}{c|}{0.155}          & \multicolumn{1}{c|}{0.220}          & 0.266          & \multicolumn{1}{c|}{0.066}          & \multicolumn{1}{c|}{0.100}          & 0.128          & \multicolumn{1}{c|}{0.043}          & \multicolumn{1}{c|}{0.073}          & 0.097          & \multicolumn{1}{c|}{0.244}          & \multicolumn{1}{c|}{0.315}          & 0.351          \\ \hline
\end{tabular}
}}
\caption{The sufficiency of explanation methods for the different types of fakes (DF, F2F, FS, NT) in the FaceForensics++ dataset, after performing adversarial attacks on the image regions corresponding to the top-1, top-2 and top-3 scoring segments based on the different explanation methods. Best (highest) scores in bold and second best scores underlined.}
\label{tab:sufficiency}
\end{table*}

The accuracy of the employed deepfake detector for the different types of fakes, on the original set of images (second row) and the generated variants of them after performing adversarial attacks on the image regions corresponding to the top-1, top-2 and top-3 scoring segments according to the different explanation methods, is presented in \cref{tab:accuracy}. These results show that LIME$_{adv}$ is the top-performing method for all types of fakes and experimental settings, as it is associated with the largest drop in detection accuracy. Regarding the remaining methods, SHAP$_{adv}$ seems to be the most competitive one, surpassing in most cases the performance of SOBOL$_{adv}$, while RISE$_{adv}$ exhibits the weakest performance. Moreover, a pairwise comparison between the original implementation and the modified version of each explanation method, shows that the proposed use of adversarially-generated images as perturbation masks, leads to lower (better) accuracy scores in most cases. As depicted in \cref{fig:accuracy_graphs}, LIME$_{adv}$ exhibits consistently better performance than LIME, leading to further drop in detection accuracy (by $10.5\%$ on average), which can be up to $18.5\%$ in some cases. Similar observations can be made by comparing SHAP with SHAP$_{adv}$ and SOBOL with SOBOL$_{adv}$. Only in the case of RISE we notice some mixed results; though, the observed improvement in the case of DF and NT fakes is higher than the decline in the remaining classes of fakes. These findings document the positive contribution of the proposed data perturbation approach on the performance of most of the considered explanation methods.

The sufficiency scores of the explanation methods are shown in \cref{tab:sufficiency}. Once again, the modified version of LIME exhibits consistently better performance than the modified versions of the remaining methods, which are ranked as before. Moreover, it outperforms its original counterpart in all cases. Finally, pairwise comparisons between the original and modified version of each explanation method document again that the impact of the proposed perturbation approach in the deepfake explanation performance is mostly positive.

Finally, we studied the computational complexity introduced by the proposed perturbation approach, by counting the time (in sec.) and the model inferences needed for producing an explanation. The results in \cref{tab:complexity} show that our approach increases the complexity of all methods, as expected. However, focusing on the best-performing explanation method (LIME), we argue that the introduced computation overhead is balanced by the observed performance gains, and does not restrict the use of LIME$_{adv}$ for obtaining explanations during real-life deepfake detection tasks.

\subsection{Qualitative results}

Our qualitative analysis is based on four pairs of real and manipulated images of the FaceForensics++ dataset (one pair per type of fake), that are shown in the top two rows of \cref{fig:qualitative_full}, and the produced visual explanations by the different methods, that are presented in the remaining rows of \cref{fig:qualitative_full}. A comparison between LIME and LIME$_{adv}$ shows that LIME$_{adv}$: i) defines more completely the manipulated area in the case of the DF sample (including the region around the added eyeglasses), ii) spots more accurately the regions close to the eyes, mouth and chin in the case of the F2F sample (that are typically altered when transferring the expressions from a source to a target video), iii) demarcates more accurately the modified region in the case of FS (avoiding to highlight parts of the background), and iv) puts more focus on the manipulated chin of the NT sample, that was missed by the LIME-based explanation. Similar remarks can be made for the other methods, since in most cases: i) SHAP$_{adv}$ appears to produce more complete and better-focusing explanations compared to SHAP ii) SOBOL$_{adv}$ seems to define more accurately the manipulated regions than SOBOL, and iii) RISE$_{adv}$ exhibits higher sufficiency in demarcating the manipulated regions, compared to RISE. These observations demonstrate the improved performance of the modified explanation methods and document once more the positive impact of the proposed perturbation approach.

\begin{table}[t]
\centering
{\small{
\begin{tabular}{|l|cc|cc|}
\hline
      & \multicolumn{2}{c|}{\begin{tabular}[c]{@{}c@{}}Computing time \\ per image (in sec.)\end{tabular}} & \multicolumn{2}{c|}{\begin{tabular}[c]{@{}c@{}}Number of \\inferences per image\end{tabular}} \\ \hline
      & \multicolumn{1}{c|}{Original}                       & Modified                      & \multicolumn{1}{c|}{Original}                      & Modified                     \\ \hline
LIME  & \multicolumn{1}{c|}{4.4}                            & 20.8                             & \multicolumn{1}{c|}{200}                           & 2081                            \\ \hline
SHAP  & \multicolumn{1}{c|}{2.6}                            & 18.7                             & \multicolumn{1}{c|}{45}                            & 1928                            \\ \hline
SOBOL & \multicolumn{1}{c|}{0.4}                            & 16.9                             & \multicolumn{1}{c|}{17}                            & 1898                            \\ \hline
RISE  & \multicolumn{1}{c|}{1.8}                            & 18.8                             & \multicolumn{1}{c|}{32}                            & 1913                            \\ \hline
\end{tabular}
}}
\caption{The computational complexity of the original and modified explanation methods.}
\vspace{-2mm}
\label{tab:complexity}
\end{table}

\begingroup
\setlength{\tabcolsep}{1pt}
\renewcommand{\arraystretch}{0.18}
\begin{figure*}[t!]
    \centering
    \begin{tabular}{m{0.12\textwidth} m{0.12\textwidth} m{0.12\textwidth} m{0.12\textwidth} m{0.12\textwidth}}
        Original \newline Images &
        \includegraphics[width=\linewidth]{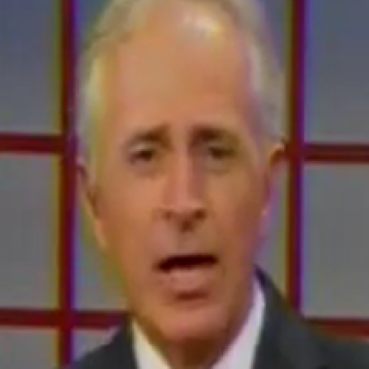} &
        \includegraphics[width=\linewidth]{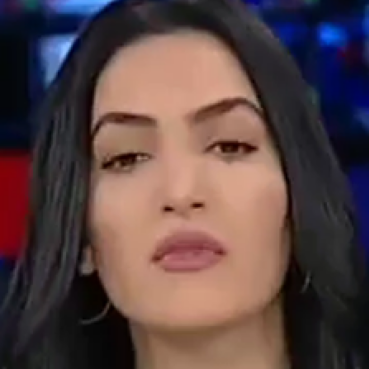} &
        \includegraphics[width=\linewidth]{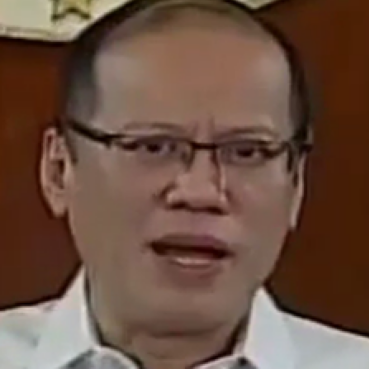} &
        \includegraphics[width=\linewidth]{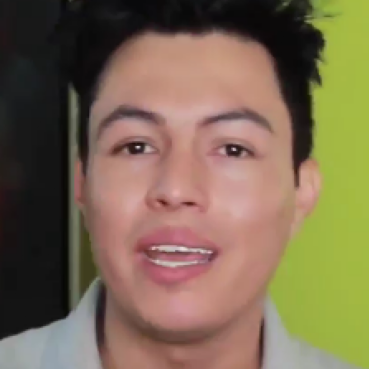} \\
    \end{tabular}
    \begin{tabular}{m{0.12\textwidth} m{0.12\textwidth} m{0.12\textwidth} m{0.12\textwidth} m{0.12\textwidth}}
        Manipulated \newline Images &
        \includegraphics[width=\linewidth]{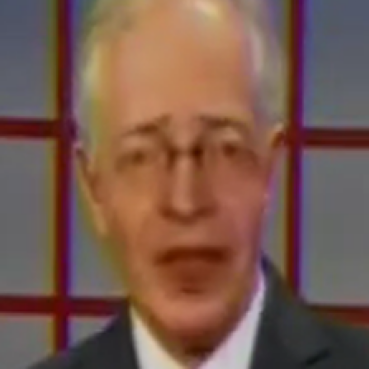} &
        \includegraphics[width=\linewidth]{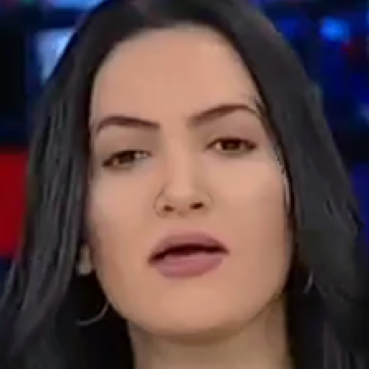} &
        \includegraphics[width=\linewidth]{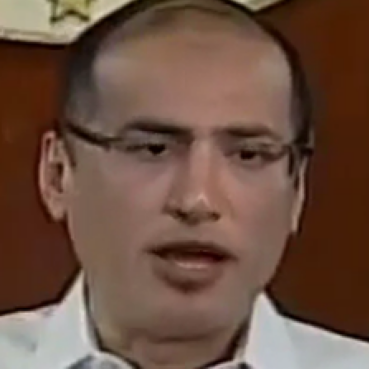} &
        \includegraphics[width=\linewidth]{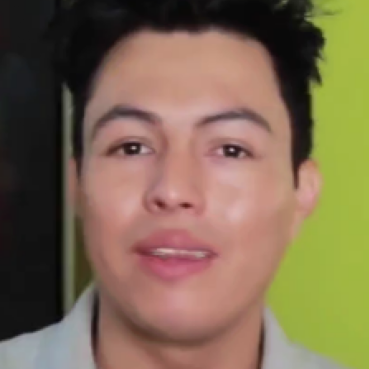} \\
    \end{tabular}

    \begin{tabular}{m{0.12\textwidth} m{0.12\textwidth} m{0.12\textwidth} m{0.12\textwidth} m{0.12\textwidth}}
        & \centering DF  
        & \centering F2F  
        & \centering FS  
        & \centering NT 
    \end{tabular}

    \begin{tabular}{m{0.12\textwidth} m{0.12\textwidth} m{0.12\textwidth} m{0.12\textwidth} m{0.12\textwidth}}
        LIME &
        \includegraphics[width=\linewidth]{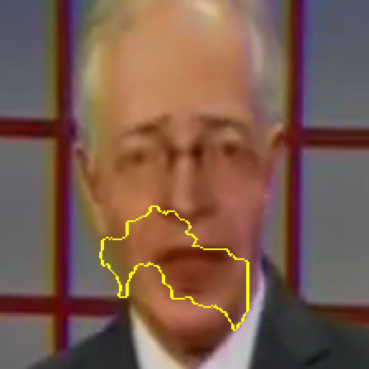} &
        \includegraphics[width=\linewidth]{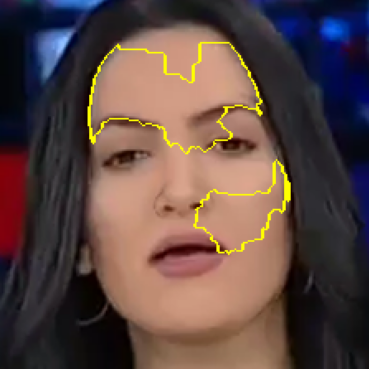} &
        \includegraphics[width=\linewidth]{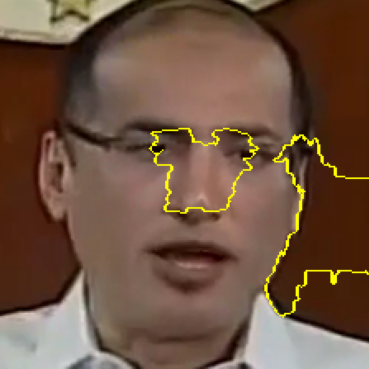} &
        \includegraphics[width=\linewidth]{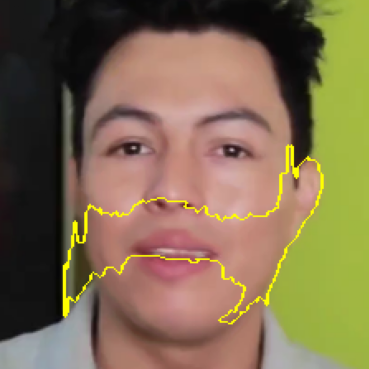} \\
    \end{tabular}

    \begin{tabular}{m{0.12\textwidth} m{0.12\textwidth} m{0.12\textwidth} m{0.12\textwidth} m{0.12\textwidth}}
        LIME$_{adv}$ &
        \includegraphics[width=\linewidth]{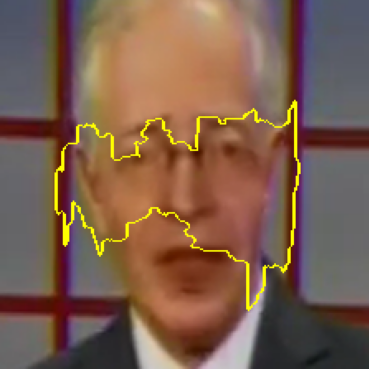} &
        \includegraphics[width=\linewidth]{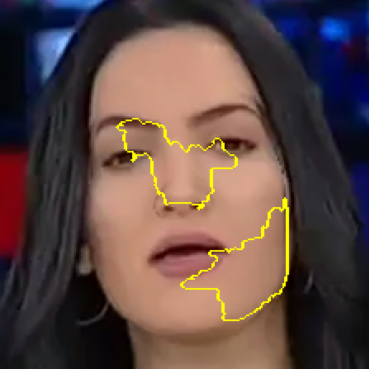} &
        \includegraphics[width=\linewidth]{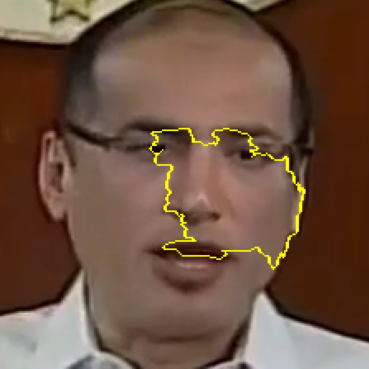} &
        \includegraphics[width=\linewidth]{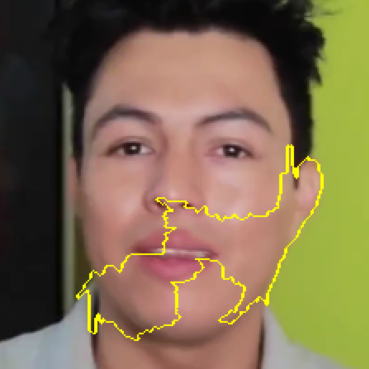} \\
    \end{tabular}

    \begin{tabular}{m{0.12\textwidth} m{0.12\textwidth} m{0.12\textwidth} m{0.12\textwidth} m{0.12\textwidth}}
        SHAP &
        \includegraphics[width=\linewidth]{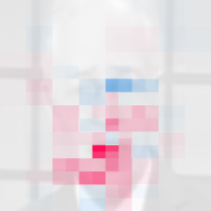} &
        \includegraphics[width=\linewidth]{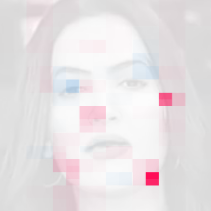} &
        \includegraphics[width=\linewidth]{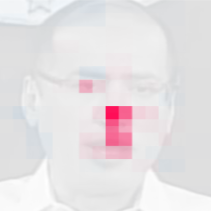} &
        \includegraphics[width=\linewidth]{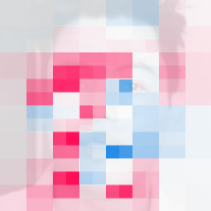} \\
    \end{tabular}

    \begin{tabular}{m{0.12\textwidth} m{0.12\textwidth} m{0.12\textwidth} m{0.12\textwidth} m{0.12\textwidth}}
        SHAP$_{adv}$ &
        \includegraphics[width=\linewidth]{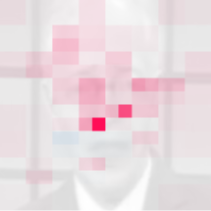} &
        \includegraphics[width=\linewidth]{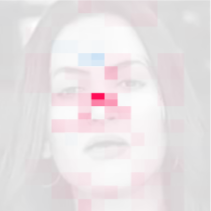} &
        \includegraphics[width=\linewidth]{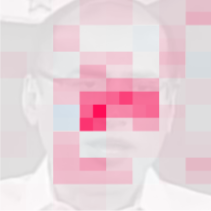} &
        \includegraphics[width=\linewidth]{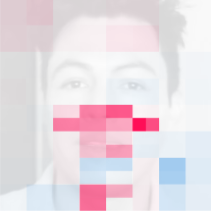} \\
    \end{tabular}

        \begin{tabular}{m{0.12\textwidth} m{0.12\textwidth} m{0.12\textwidth} m{0.12\textwidth} m{0.12\textwidth}}
        SOBOL &
        \includegraphics[width=\linewidth]{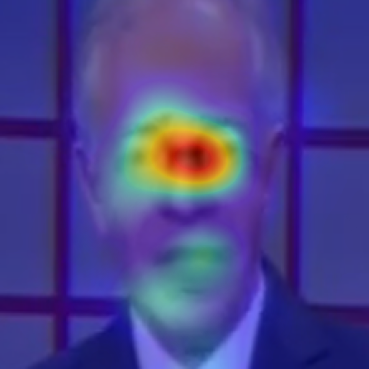} &
        \includegraphics[width=\linewidth]{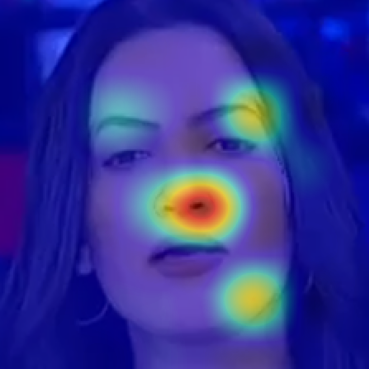} &
        \includegraphics[width=\linewidth]{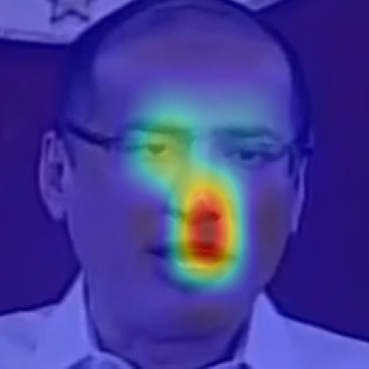} &
        \includegraphics[width=\linewidth]{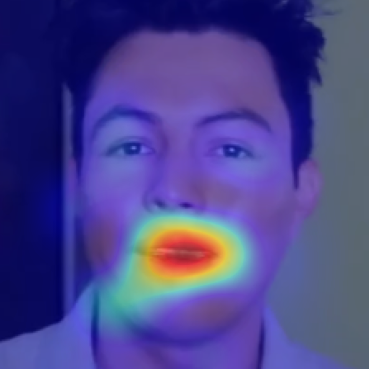} \\
    \end{tabular}

    \begin{tabular}{m{0.12\textwidth} m{0.12\textwidth} m{0.12\textwidth} m{0.12\textwidth} m{0.12\textwidth}}
        SOBOL$_{adv}$ &
        \includegraphics[width=\linewidth]{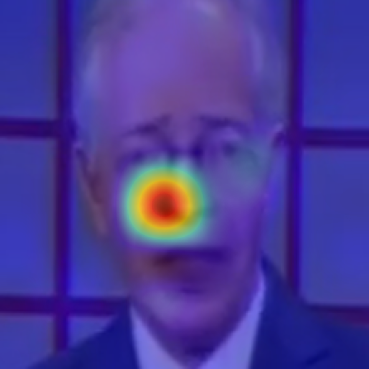} &
        \includegraphics[width=\linewidth]{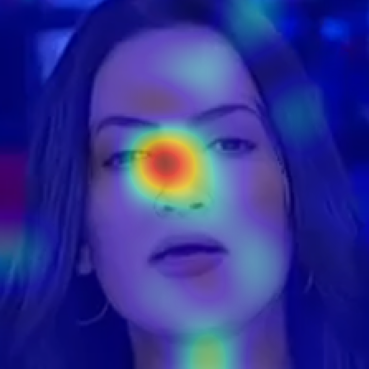} &
        \includegraphics[width=\linewidth]{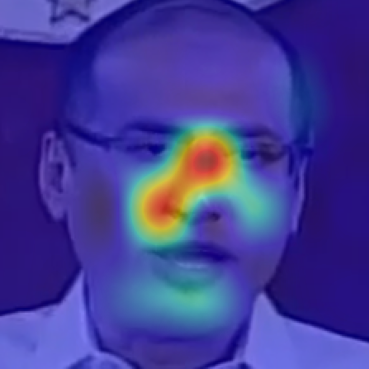} &
        \includegraphics[width=\linewidth]{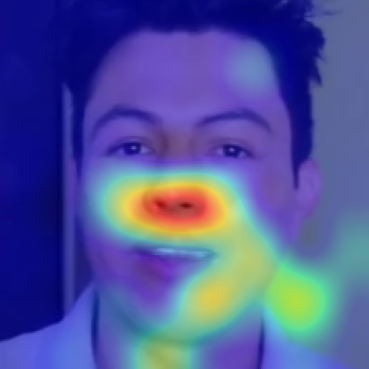} \\
    \end{tabular}

    \begin{tabular}{m{0.12\textwidth} m{0.12\textwidth} m{0.12\textwidth} m{0.12\textwidth} m{0.12\textwidth}}
        RISE &
        \includegraphics[width=\linewidth]{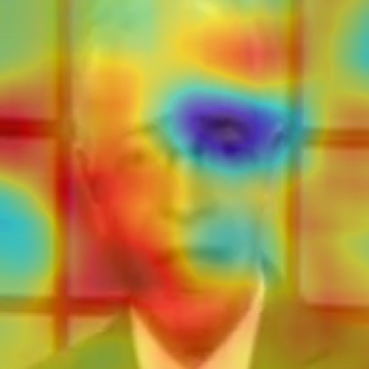} &
        \includegraphics[width=\linewidth]{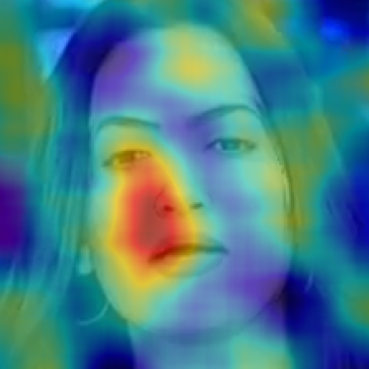} &
        \includegraphics[width=\linewidth]{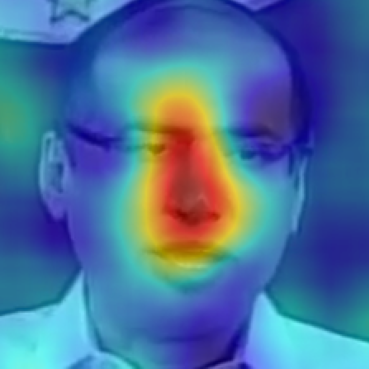} &
        \includegraphics[width=\linewidth]{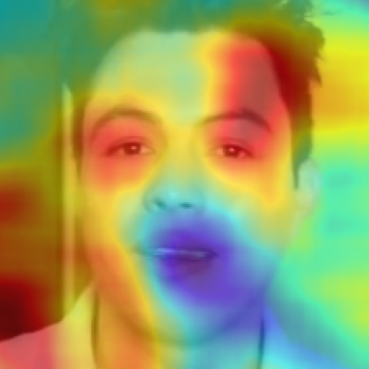} \\
    \end{tabular}

    \begin{tabular}{m{0.12\textwidth} m{0.12\textwidth} m{0.12\textwidth} m{0.12\textwidth} m{0.12\textwidth}}
        RISE$_{adv}$ &
        \includegraphics[width=\linewidth]{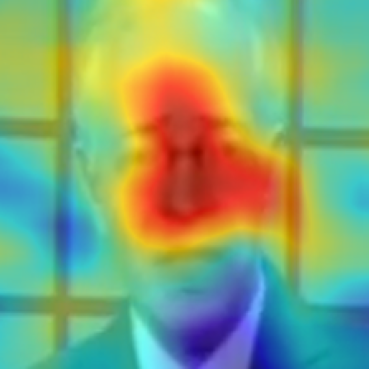} &
        \includegraphics[width=\linewidth]{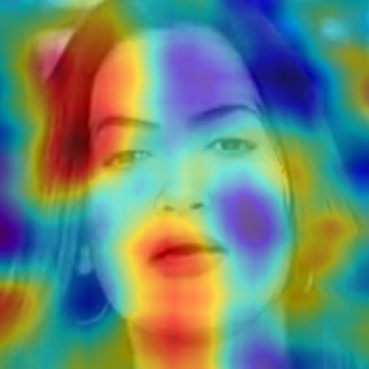} &
        \includegraphics[width=\linewidth]{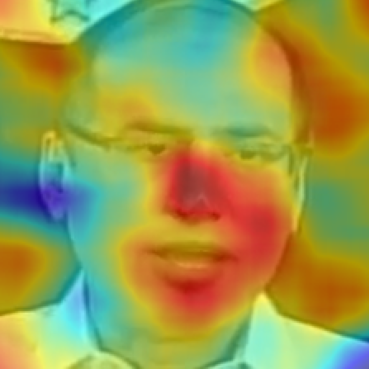} &
        \includegraphics[width=\linewidth]{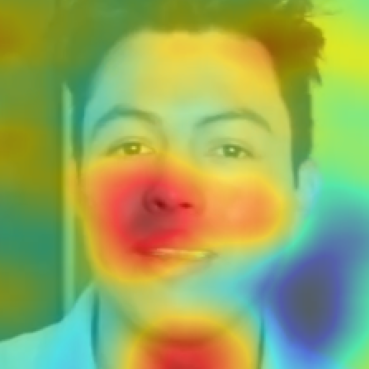} \\
    \end{tabular}
    
     \caption{The obtained explanations per type of manipulation (displayed using the default visualization format of each explanation method).}
     \label{fig:qualitative_full}
\end{figure*}
\endgroup

\section{Conclusions}

In this work, we presented our idea for improving the performance of perturbation-based methods when explaining deepfake detectors. This idea relies on the use of adversarially-generated samples of the input deepfake images, to form perturbation masks for inferring the importance of input features. We integrated the proposed perturbation approach in four SOTA explanation methods and evaluated the performance of the resulting modified methods on a benchmarking dataset. The conducted quantitative and qualitative analysis documented the gains in the performance of most of these methods, that can support the production of more accurately defined visual explanations.

{\small

}

\end{document}